%% file: neurips_2025.tex
\documentclass{article}

\usepackage[preprint]{neurips_2025}

\usepackage[utf8]{inputenc} 
\usepackage[T1]{fontenc}    
\usepackage{hyperref}       
\usepackage{url}            
\usepackage{booktabs}       
\usepackage{amsfonts}       
\usepackage{nicefrac}       
\usepackage{microtype}      
\usepackage{xcolor}         

\usepackage{multirow}
\usepackage{graphicx}
\usepackage{adjustbox}
\usepackage{float}
\usepackage{pifont}
\usepackage{svg}
\usepackage{algorithm}
\usepackage{algpseudocode}
\usepackage{tikz}
\usepackage{tabularx}
\usepackage{xcolor}
\usepackage{amsmath}
\usepackage{wrapfig}
\usepackage{subcaption}    
\usepackage{wrapfig}
\usepackage{enumitem}

\definecolor{blue}{RGB}{34,34,225}
\definecolor{myGreen}{RGB}{34, 139, 34}
\definecolor{red}{RGB}{225, 34, 34}

\newcommand{\proposed}{\texttt{DreamPose3D}}

\title{DreamPose3D: Hallucinative Diffusion with Prompt Learning for 3D Human Pose Estimation}

\author{%
	Jerrin Bright$^{1,2}$, 
	Yuhao Chen$^{1,2}$, 
	John S. Zelek$^{1,2}$,\\
	$^{1}$Vision and Image Processing Lab, 
	$^{2}$University of Waterloo
}

\begin{document}

\maketitle

\begin{abstract}
    Accurate 3D human pose estimation remains a critical yet unresolved challenge, requiring both temporal coherence across frames and fine-grained modeling of joint relationships. However, most existing methods rely solely on geometric cues and predict each 3D pose independently, which limits their ability to resolve ambiguous motions and generalize to real-world scenarios. Inspired by how humans understand and anticipate motion, we introduce {\proposed}, a diffusion-based framework that combines action-aware reasoning with temporal imagination for 3D pose estimation. {\proposed} dynamically conditions the denoising process using task-relevant action prompts extracted from 2D pose sequences, capturing high-level intent. To model the structural relationships between joints effectively, we introduce a representation encoder that incorporates kinematic joint affinity into the attention mechanism. Finally, a hallucinative pose decoder predicts temporally coherent 3D pose sequences during training, simulating how humans mentally reconstruct motion trajectories to resolve ambiguity in perception. Extensive experiments on benchmarked Human3.6M and MPI-3DHP datasets demonstrate state-of-the-art performance across all metrics. To further validate \textbf{\proposed}'s robustness, we tested it on a broadcast baseball dataset, where it demonstrated strong performance despite ambiguous and noisy 2D inputs, effectively handling temporal consistency and intent-driven motion variations.
\end{abstract}

\input{sec/1_intro}

\input{sec/2_related_works}
\input{sec/3_method}
\input{sec/4_experiments}
\input{sec/5_conclusion}
\input{sec/X_suppl}

\newpage
\bibliographystyle{plainnat}
\bibliography{main}

\end{document}

%% file: sec/1_intro.tex
\section{Introduction}\label{sec:intro}

Given monocular 2D images or videos, 3D Human Pose Estimation (3D HPE) aims to predict the positions of human body joints in 3D space. It is crucial in applications such as augmented and virtual reality \cite{mehta2017vnect}, sports analysis \cite{bright2024pitchernet, bright2023mitigating, rematas2018soccer}, self-driving \cite{driving1, driving2, driving3}, and robotics \cite{robot_3dpe, robot_3dpe2}. Traditional monocular 3D HPE \cite{mixste, videopose3d, actionprompt, pstmo, motionbert2022} focuses on reconstructing plausible 3D poses from 2D projections \cite{hrnet, xu2022vitpose, cpn}. However, despite being conditioned on a 2D pose sequence, these approaches typically predict 3D poses on a per-frame basis, thereby neglecting longer-term motion cues that could improve temporal coherence and robustness.


Recent diffusion-based methods \cite{d3dp, diffpose_cvpr, diffupose, diffpose_iccv, feng2023diffpose, zhou2023diff3dhpe} have explored generative modeling for 3D HPE, sampling plausible 3D skeletons conditioned on local 2D pose sequences. Despite their probabilistic formulation, these techniques limit their predictions to a single target frame.
Moreover, these models rely solely on geometric information from 2D poses, without considering the higher-level context or intent behind the motion. We refer to this limitation as \textit{intent ambiguity}. This can lead to uncertainty when interpreting motion patterns, especially when different actions exhibit similar joint movements over short windows, such as waving vs. throwing, or jumping vs. stumbling. This limits their generalizability in scenarios where visual cues may be noisy or corrupted. In such cases, semantic context (i.e., action-level priors) can help guide pose reconstruction more effectively.


\begin{figure*}[t]
  \centering

  \begin{minipage}[t]{0.49\textwidth}
    \centering
    \includegraphics[width=\linewidth]{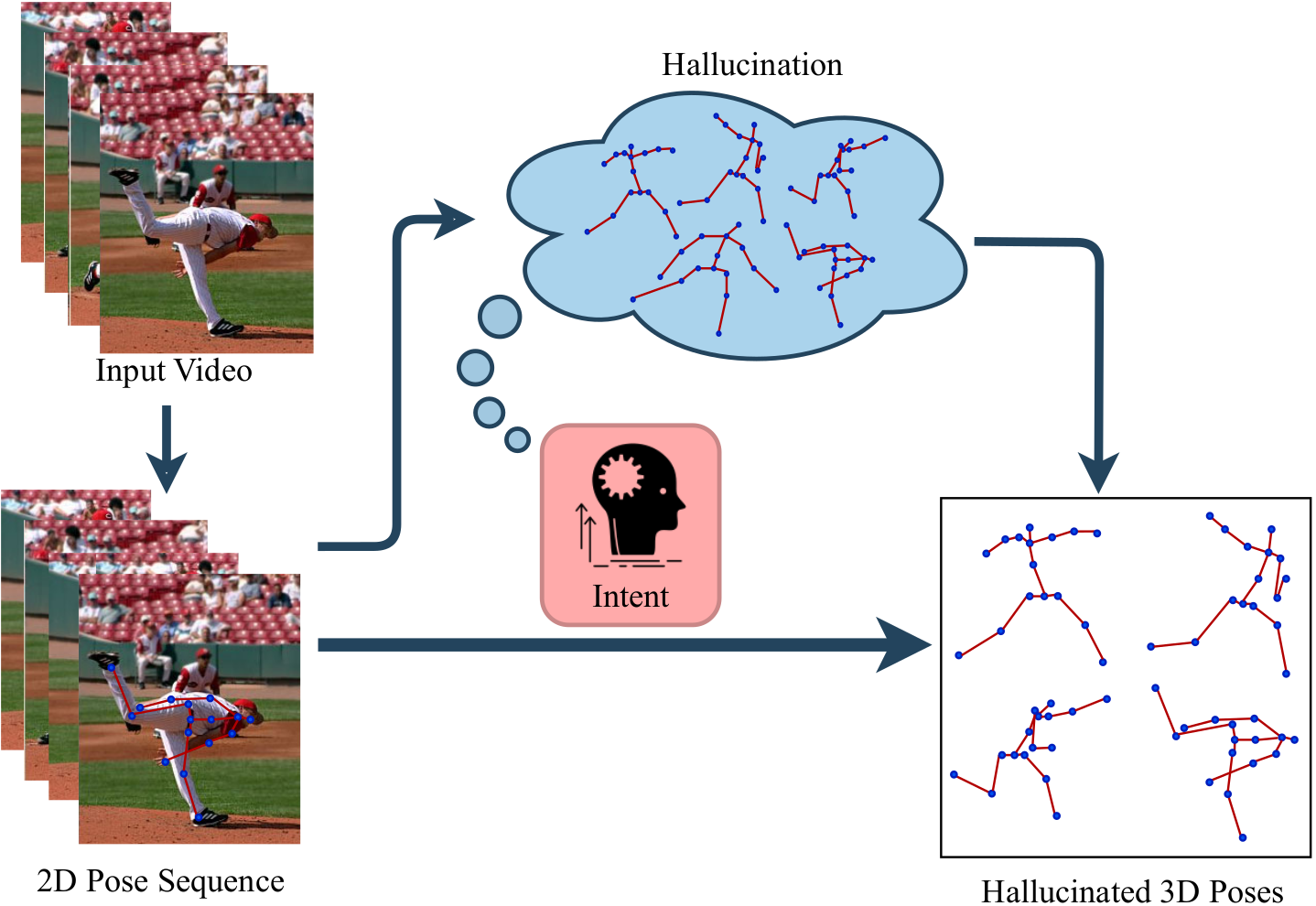}
    \subcaption{High-level overview of {\proposed} framework.}
    \label{fig:teaser-a}
  \end{minipage}
  \hfill
  \begin{minipage}[t]{0.49\textwidth}
    \centering
    \includegraphics[width=\linewidth]{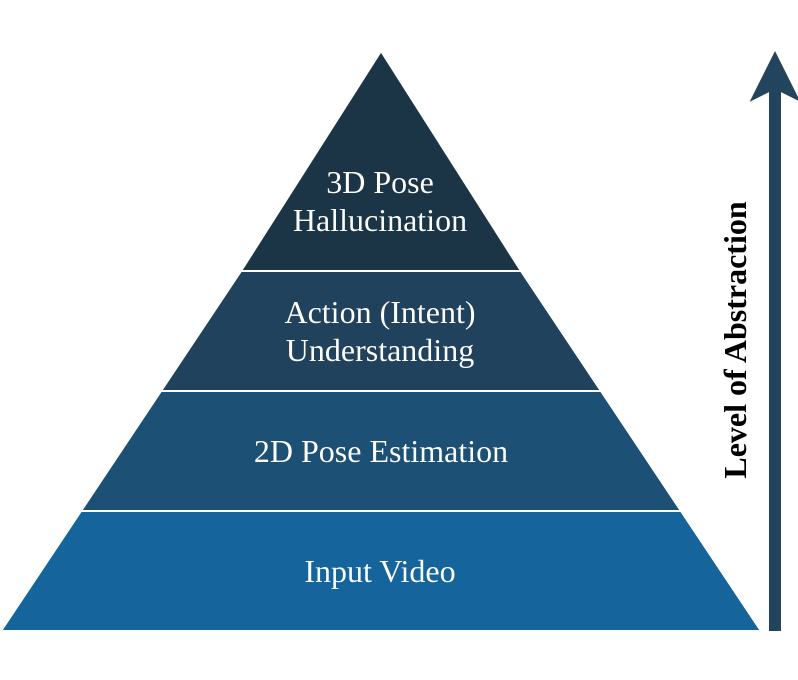}
    \subcaption{Illustration of intent-driven motion perception.}
    \label{fig:intent}
  \end{minipage}
  \caption{\textbf{Overview and Effectiveness of \proposed.} (a) A conceptual overview of the proposed framework, which mimics human-like reasoning of motion. (b) Qualitative impact of intent and hallucination modules, showing performance with and without each component.}
  \label{fig:teaser}
\end{figure*}

Prior work in cognitive science has shown that human perception of motion involves both intent recognition and mental simulation. A recent study on kinematic priming \cite{scaliti2023kinematic} demonstrated that prior knowledge of a person’s intent, such as drinking versus pouring, significantly alters how observers interpret subsequent movement kinematics. This suggests that intent strongly shapes motion understanding, especially under ambiguous visual input. The hierarchy illustrated in Figure \ref{fig:intent} reflects this cognitive sequence, moving from low-level visual cues to high-level motion simulation. Motivated by these insights, we ask: \textit{Is it possible to build a model that mimics the way humans instinctively reason about motion by identifying intent and imagining plausible motion that aligns with it?}


To mimic this human-like process of reasoning about motion, we reformulate 3D HPE as a \textbf{intent-driven motion imagination task}. Instead of predicting a single 3D pose at a time, our method imagines a coherent sequence of 3D poses that span the past, present, and future, given an input window of 2D poses. This allows the model to understand not just the 3D pose at the current frame, but how it evolves explicitly over time. We further condition the generation using action-aware prompts, encoded using a pre-trained Vision-Language Model (VLM). These prompts act as the high-level intent, helping the model generate contextually meaningful poses.

In this work, we propose {\proposed}, a prompt-driven hallucinative diffusion model for 3D human pose estimation (Figure \ref{fig:teaser-a}). {\proposed} first employs a transformer network to learn motion representations from 2D pose sequences to predict the action prompt. Then, using a pre-trained VLM (CLIP \cite{radford2021learning}), it encodes context embeddings based on the learned action prompt. These embeddings condition a diffusion model that denoises a Gaussian distribution using spatial and temporal multi-head attention. We further modify the attention mechanism by infusing it with the global kinematic joint relationships, ensuring consistent pose structure across frames. The output tokens of the diffusion model are then decoded using a pose hallucinator that predicts a sequence of \textit{hallucinatory} 3D poses, enforcing temporal consistency by regularizing motion dynamics. 


Our contributions can be summarized as follows:


\begin{itemize}
    \item We propose \textbf{\proposed}, a prompt-driven hallucinative diffusion model that addresses temporal inconsistencies and intent ambiguity in 3D human pose estimation.
    
    \item We introduce a lightweight pose hallucinator that generates temporally coherent 3D pose sequences, improving motion continuity over time.
    
    \item Our method achieves state-of-the-art performance on both Human3.6M and MPI-INF-3DHP datasets. Additional experiments on the MLBPitchDB dataset highlight its robustness in scenarios with noisy or corrupted 2D pose inputs.
\end{itemize}

%% file: sec/2_related_works.tex
\section{Related Work}\label{sec:related-work}

\subsection{3D Human Pose Estimation} 

Traditional 3D human pose estimation methods decompose the task by first extracting 2D poses from input images or videos, then reconstructing the 3D pose from these 2D pose sequences \cite{mixste, videopose3d, poseformerv2, motionbert2022}. These approaches predict a single, most likely 3D pose for each 2D observation/ sequence. Recently, however, multihypothesis approaches have emerged, generating a set of plausible 3D poses for a given timestamp $f$ from a given 2D pose sequence centered around $f$ \cite{d3dp, diffpose_iccv, mhformer}. These approaches aim to better capture pose variability by combining multiple plausible 3D predictions using conditioning or averaging techniques.

By contrast, we predict a set of $n$ sequential 3D poses from the given 2D pose sequence. With our simple yet novel hallucinating approach, we are able to reflect the full dynamic progression of poses.   


\subsection{Diffusion Models}

Diffusion models \cite{ddpm} have become state-of-the-art in various generative tasks. The diffusion model gradually removes noise from an initialized Gaussian distribution, generating outputs that align with the target distribution. They have demonstrated superior performance for various computer vision tasks including test-to-image synthesis \cite{t2i1, t2i2}, super-resolution \cite{superres1, superres2}, image inpainting \cite{inpaint1, inpaint2} and segmentation \cite{seg1, seg2}. Recent works on 3D HPE \cite{d3dp, diffpose_cvpr, diffpose_iccv, diffupose} have focused on using diffusion models, with the aim of naturally handling the indeterminacy and uncertainty in the observation. Diffusion models for 3D HPE offer various advantages: (1) Plausible human poses against noise and occlusions \cite{Zhou_2023_ICCV}; (2) Does not suffer from phenomena like posterior collapse, vanishing gradients, or training instabilities \cite{diffpose_iccv}; (3) Captures fine-grained dynamics with fidelity even during inherent ambiguities in representation \cite{Zhou_2023_ICCV, feng2023diffpose}.


Thus, to address the inherent indeterminacy and uncertainty in 3D HPE, we leverage the strengths of diffusion models, which are well-suited for handling such challenges. Thus, we formulate our 3D HPE task as a reverse diffusion process, enabling {\proposed} to generate accurate 3D pose reconstructions by effectively managing the ambiguous and uncertain nature of pose predictions.




\subsection{Prompt Learning}

Text prompts have emerged as a valuable tool for guiding various vision tasks, including pose estimation. Several recent works \cite{clamp, lamp, hu2024animal, tokenclipose} have used text prompts to estimate 2D poses of hands, humans, and animals, respectively. More recently, MDM \cite{tevet2023human} utilized text prompts to generate 3D pose sequences using diffusion models, and FinePOSE \cite{xu2024finepose} introduced a part-aware learning mechanism that encodes information about action classes and coarse- and fine-grained human parts.

Inspired by these works, we propose the Action Prompt Learning (APL) block, which introduces learnable prompt embeddings to infer high-level motion intent and guide the 3D pose generation process. Unlike previous approaches \cite{xu2024finepose, clamp} that rely on explicit action class labels, {\proposed} distinguishes itself by \textit{directly learning the action} by encoding the motion through a transformer backbone. This enables the model to operate in scenarios where action labels are unavailable. 

%% file: sec/3_method.tex
\begin{figure*}[t]
  \centering
  \begin{tikzpicture}
    \node at (0,0) {\includegraphics[width=\linewidth]{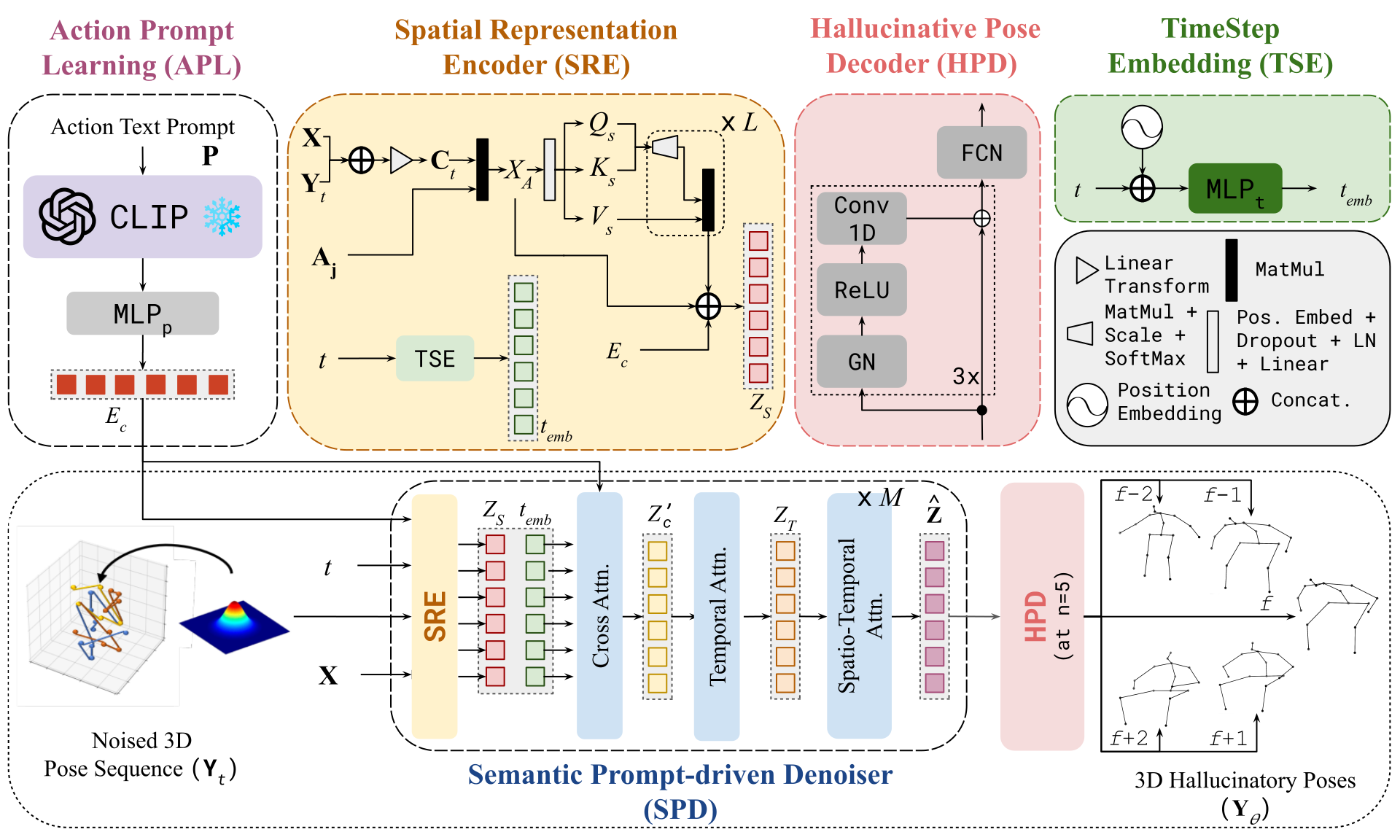}};
  \end{tikzpicture}
   \vspace{-20px}
  \caption{\textbf{Overview of {\proposed} model architecture.} Given an input 2D pose sequence $\textbf{X}$, {\proposed} predicts temporally coherent 3D poses through the SPD block, which performs diffusion-based denoising. SPD is conditioned on a context embedding, inferred from the 2D sequence via the APL block. The HPD block complements SPD by generating auxiliary 3D hallucinatory poses to reinforce temporal consistency across frames.}
  \label{fig:overview}
\end{figure*}

\section{Method} \label{sec:method}


Given an input 2D skeleton sequence $\textbf{X} \in \mathbb{R}^{N \times J \times 2}$, defined by $N$ frames with $J$ joints for each skeleton, the goal of our approach is to predict the corresponding 3D pose sequence $\textbf{Y}_{\theta} \in \mathbb{R}^{N \times J \times 3}$. To address the challenges of temporal coherence and intent ambiguity in 3D HPE, {\proposed} incorporates three core modules. First, the Action Prompt Learning (APL) block generates action-aware embeddings from the input 2D pose sequence, encoding intent for realistic pose transitions. Next, a diffusion model utilizes the proposed Semantic Prompt-driven Denoiser (SPD) in the reverse diffusion process, improving denoising performance with spatially and temporally informed embeddings. Finally, a Hallucinative Pose Decoder (HPD) module predicts $n$ 3D hallucinatory poses, ensuring smooth, coherent transitions across frames. Figure \ref{fig:overview} shows the network architecture of {\proposed} for 2D-to-3D lifting of the skeleton sequences.

\subsection{Action Prompt Learning}

Existing works often rely on \textit{manual text prompts} along with input images or pose sequences to guide their models toward a specific task \cite{clamp, xu2024finepose, tokenclipose}. To address the limitations of manual prompts, which may not always be available, we propose the APL block that automatically generates task-relevant prompts from the 2D pose sequence (\textbf{X}) and encodes context-aware information.     


The APL block consists of two main components: an intent classifier and a VLM. Given the input 2D pose sequence, the intent classifier encodes the motion representation using a transformer-based encoder \cite{motionbert2022} denoted as $\textbf{E}$. This motion representation is then passed through a lightweight decoder ($\phi$) composed of global average pooling, deconvolution layers, and an MLP with a single hidden layer to generate a context-aware text prompt ($\textbf{P}$), as shown in Equation \eqref{eq:prompt}.
\vspace{-2px}
\begin{equation}
    \mathbf{P} = \phi( \textbf{E}(\textbf{X}))
    \label{eq:prompt}
\end{equation}

The generated prompt, which captures the semantics of the observed 2D motion, is then tokenized using the standard CLIP tokenizer ($\text{t}_k$), and fed as input to a VLM (CLIP \cite{radford2021learning}). Specifically, CLIP operates with a fixed token sequence length of 77 tokens, which we structurally partition into two learnable prompt segments: 40 tokens for subject-related context (i.e., template: "\textit{a person}") and 37 for action-related context. 

Explicitly regularizing the prompt guides the model towards better convergence and stability while also offering flexibility by allowing the use of custom or predicted prompts. The CLIP output embeddings are then passed through an MLP block ($\text{MLP}_p$) to obtain the context embeddings ($E_{c}$), as formulated in Equation \eqref{eq:apl}.
\vspace{-2px}
\begin{equation}
    \begin{split}
    E_{c} = \text{MLP}_p[\text{CLIP}(\text{t}_k(\textbf{P}))] 
    \label{eq:apl}
\end{split}
\end{equation}


\subsection{Semantic Prompt-driven Denoiser}

The SPD block is the key component in the reverse process of {\proposed}, designed to output coherent 3D pose tokens ($\hat{\textbf{Z}}$) by leveraging spatial and temporal context. As the denoiser network $\mathcal{D}$, the SPD aims to progressively refine noisy 3D pose sequences by conditioning on multiple inputs: the noised 3D pose sequence ($\textbf{Y}_t$), the input 2D pose sequence $\textbf{X}$, context embeddings $E_{c}$, and time $t$. The resulting output is a set of denoised pose tokens ($\hat{\textbf{Z}}$), which effectively incorporate human joint kinematics across spatial and temporal dimensions, formulated as shown in Equation \eqref{eq:z_hat}.
\vspace{-2px}
\begin{equation}
    \hat{\textbf{Z}} = \mathcal{D}(\textbf{Y}_t, \textbf{X}, E_{c}, t)
    \label{eq:z_hat}
\end{equation}

\noindent \textbf{Spatial Representation Encoder (SRE).} The SRE encodes spatial dependencies by modeling joint affinities, allowing the SPD to learn local and global relationships between joints in each frame. Initially, the 2D pose $\textbf{X}$ and the noisy 3D pose $\textbf{Y}_t$ are concatenated to form the pose tokens ($\textbf{C}_t$), representing joint configurations across frames. 

Inspired by affinity mechanisms \cite{ktpformer}, which effectively capture joint dependencies in pose estimation, we construct a joint affinity matrix ($\textbf{A}_\textbf{j}$) to incorporate local and global joint interactions, which are critical for ensuring joint alignment. The joint affinity matrix $\textbf{A}_\textbf{j}$ is defined as:
\vspace{-2px}
\begin{equation}
    \textbf{A}_\textbf{j} = \dfrac{(A_L + A_G) + (A_L + A_G)^T}{2} \label{eq:aff_matrix}
\end{equation}
 
\noindent where $A_L$ represents local joint affinities for proximal joints (hand-crafted), and $A_G$ captures global affinities for joints at greater spatial distances (learnable). This symmetric affinity matrix enables the model to understand each joint's structural dependencies with respect to neighboring and distant joints. To integrate the spatial relationships from $\textbf{A}_\textbf{j}$ with the pose tokens $\textbf{C}_t$, we apply a series of multi-head attention layers. First, the affinity-modulated pose tokens $X_A = \textbf{C}_t \cdot \textbf{A}_\textbf{j}$ are linearly transformed to obtain queries ($Q_s$), keys ($K_s$), and values ($V_s$). These representations are then processed through spatial multi-head attention with $L$ heads.
The multi-head attention output is subsequently combined with the affinity-modulated tokens ($X_A$), along with the context embeddings $E_{c}$, which acts as a residual connection to enhance spatial consistency in the final pose tokens ($Z_S$). The overall output of the SRE, incorporating the spatial multi-head attention, can be expressed as:
\vspace{-2px}
\begin{equation}
    Z_S = {MultiHead}(Q_s,K_s,V_s) + X_A + E_{c}
\end{equation}

\paragraph{TimeStep Embedding.} Following DDPMs, a sinusoidal function encodes the timestep $t$. This encoded timestep is then processed through an MLP block ($\text{MLP}_t$), which consists of linear layers and GELU activation functions. The output of $\text{MLP}_t$, denoted as $t_{emb}$, is added to the SRE block and subsequently fed as an additional input to the temporal attention block. This ensures the SPD block can effectively handle 3D noisy poses at various timesteps.

\paragraph{Cross Attention.} To incorporate high-level context, we implement a cross-attention mechanism using the output tokens $Z_S$ from the SRE block and the context embeddings $E_{c}$. Here, the query, key, and value are defined as $Q_c = W_q Z_S$, $K_c = W_k E_{c}$, and $V_c = W_v E_{c}$, respectively, where $W_q$, $W_k$, and $W_v$ are the respective parameter matrices. The resulting output tokens from the cross-attention block are denoted as $Z_c$.
Additionally, we add the $t_{emb}$ to the output cross-attention tokens $Z_c$ to optimize based on the timestep context, denoted as $Z'_c = Z_c + t_{emb}$. Next, the concatenated tokens $Z'_c$ are passed through a temporal attention block to model inter-frame relationships between different poses, resulting in the output token representation ($Z_T$). Finally, a spatiotemporal encoder is applied to extract fine-grained and rich output representation, denoted as $\hat{\textbf{Z}}$. The spatio-temporal encoder consists of $M$ stacks of alternating spatial and temporal transformers. 

\subsection{Hallucinative Pose Decoder}

Current diffusion-based approaches \cite{xu2024finepose, d3dp} typically produce multiple plausible pose hypotheses for each frame, even when given a sequence of 2D observations. These methods often rely on additional processing to merge or refine the outputs to achieve temporal consistency. However, this probabilistic formulation lacks an inherent understanding of motion continuity across a sequence, leading to potential inconsistencies in frame-to-frame transitions.

Thus, we introduce the HPD decoder, which leverages the temporal information from the output tokens, $\hat{\textbf{Z}}$, to generate a coherent set of 3D poses termed as 3D hallucinatory poses. By predicting the hallucinatory poses across a set of consecutive frames, $\{f - \frac{n-1}{2}, \dots, f, \dots, f + \frac{n-1}{2}\}$, rather than as isolated frames, the HPD decoder inherently captures temporal dependencies, resulting in smoother and more realistic 3D pose transitions. This approach eliminates the need for further post-processing, providing a streamlined and temporally coherent output. When $n=5$, the HPD decoder as shown in Figure \ref{fig:overview}, generates poses over the sequence $\{f - 2, \dots, f + 2\}$ centered around $f$.


\subsection{Objective Functions}

The objective of {\proposed} is to minimize the error between the predicted and groundtruth poses by ensuring joint consistency through the hallucinatory poses, bone length, and action prompt regularization. The primary loss term, $\mathcal{L}_{3D}^f$ defined in Equation \eqref{eq:3dloss}, represents the $L1$ loss for $n$ 3D hallucinatory poses. 
\vspace{-2px}
\begin{equation}
    \mathcal{L}'_{3D} = \sum_{k = -\frac{n-1}{2}}^{\frac{n-1}{2}} \lambda_{3D}^{f+k} \mathcal{L}_{3D}^{f+k} \label{eq:3dloss}
\end{equation}

\noindent where $\mathcal{L}_{3D}^{f+k}$ is the 3D loss at timestamp $f+k$. To further enhance pose realism, we introduce two additional regularization terms: a cross-entropy classification loss, $\mathcal{L}_{act}$, applied to the action prompt derived from the encoder (\textbf{E}), and an $L1$ bone length regularization loss, $\mathcal{L}_{BL}$, which ensures consistent bone proportions by penalizing length discrepancies between paired joints (e.g., left and right limbs). The overall loss function of {\proposed} is given by:
\vspace{-2px}
\begin{equation}
    \mathcal{L}_{net} = \mathcal{L}'_{3D} + \lambda_{act} \mathcal{L}_{act} + \lambda_{BL} \mathcal{L}_{BL}
    \label{eq:loss}
\end{equation}

\noindent where $\lambda_{act}$, and $\lambda_{BL}$ are the weights assigned for the action class and bone length regularization terms, respectively.

\noindent \textbf{Sampling Strategy.} To enable the model to effectively learn both immediate and long-term dependencies, we adopt a two-stage sampling strategy during training. In the initial stage, we set $n=1$, focusing on accurate 3D pose prediction for the current frame only. In the second stage, we set $n>1$. Specifically, the best results were obtained at $n=3$. The weights for the 3D hallucinatory poses $\lambda_{3D}^{t+k}$ are set to be $1/(1+|k|)$ each, allowing the model to incorporate information from both past and future frames with a smaller weight to poses that are further away in time. This shift in weighting ensures the importance of hallucinated poses based on how far they are from the current frame, leading to improved accuracy in long-term pose prediction.

\noindent \textbf{Inference.} In the inference phase, {\proposed} begins by initializing the encoder $\textbf{E}$ to predict the action class and subsequently generating $E_{c}$ via the APL block. It is to note that no external input, such as action labels or prompts, is required; all contextual information is inferred directly from $\textbf{X}$. The model then performs the reverse diffusion process, sampling $N$ poses from the 2D input $\textbf{X}$ and iteratively passing them through the denoiser $\mathcal{D}$ for $K$ steps to obtain the final 3D pose sequence, represented as ${\{Y_T^h\}}_{h=1}^N$. The HPD decoder was designed mainly to improve the performance of the denoiser by regularizing its temporal understanding during training, so during inference, $n$ was always set to be 1; that is, it captures only the current frame's pose at each timestep. The resulting output for each timestep $T$ is denoted by the 3D pose $\hat{Y}^h_0$, yielding a temporally coherent sequence of $N$ poses.

%% file: sec/4_experiments.tex
\begin{table*}[t]
\caption{\textbf{Quantitative comparison with the state-of-the-art 3D HPE methods on the Human3.6M dataset.} $N$: the number of input frames. CPN, HRNet, SH: using CPN~\cite{cpn}, HRNet~\cite{hrnet}, and SH~\cite{sh} as the 2D keypoint detectors to generate the inputs. GT: using the groundtruth 2D keypoints as inputs. The best and second-best results are highlighted in \textbf{bold} and \underline{underlined} formats. Colored differences indicate the margin between the top two results.}
\adjustbox{width=\linewidth}
  {
  \setlength{\tabcolsep}{11.5pt}
  
    \begin{tabular}{lcccccccl}
    \toprule
    \multirow{2}{*}{Method} 
    & \multirow{2}{*}{$N$}  
    & \multicolumn{3}{c}{Human3.6M (DET)} 
    & \multicolumn{3}{c}{Human3.6M (GT)}
    & \multirow{2}{*}{Year} \\ 
    \cmidrule(lr){3-5} \cmidrule(lr){6-8}
    &
    & Detector & mPJPE~$\textcolor{blue}{\downarrow}$ & P-mPJPE~$\textcolor{blue}{\downarrow}$
    & Detector & mPJPE~$\textcolor{blue}{\downarrow}$ & P-mPJPE~$\textcolor{blue}{\downarrow}$\\ 
    \midrule
    VideoPose3D~\cite{videopose3d}  & 243 & CPN & 46.8 & 36.5 & GT & 37.8 &  / & \textcolor{gray}{CVPR'19} \\
    Anatomy~\cite{anatomy} & 243 & CPN & 44.1 & 35.0 & GT & 32.3 & /  & \textcolor{gray}{CSVT'21} \\
    P-STMO~\cite{pstmo} & 243 & CPN & 42.8 & 34.4 & GT & 29.3 &  / & \textcolor{gray}{ECCV'22} \\
    MixSTE~\cite{mixste} & 243 & HRNet & 39.8 & 30.6 & GT & 21.6 &  / & \textcolor{gray}{CVPR'22}\\
    MHFormer~\cite{mhformer} & 351 & CPN & 43.0 & 34.4 & GT & 30.5  & /  & \textcolor{gray}{CVPR'22}\\
    Diffpose~\cite{diffpose_iccv} & 243 & CPN & 36.9 & {28.7} & GT & 18.9 &  / & \textcolor{gray}{CVPR'23} \\
    GLA-GCN~\cite{glagcn} & 243 & CPN & 44.4 & 34.8 & GT & 21.0 & 17.6 & \textcolor{gray}{ICCV'23}\\
    ActionPrompt~\cite{actionprompt} & 243 & HRNet & 41.8 & 29.5 & GT & 22.7 & /  & \textcolor{gray}{ICME'23}\\
    NC-RetNet~\cite{ncretnet} & 243 & CPN & 40.4 & 32.5 & GT & 21.5 & / & \textcolor{gray}{ECCV'24}\\ 
    MotionBERT~\cite{motionbert2022} & 243 & SH & 37.5 & / & GT & {16.9} &  / & \textcolor{gray}{ICCV'23} \\
    D3DP~\cite{d3dp} & 243 & CPN & {35.4} & {28.7} & GT & 18.4 & / & \textcolor{gray}{ICCV'23} \\
    {KTPFormer} \cite{ktpformer} & 243 & CPN &  {33.0} & {26.2} & GT & {18.1} & / & \textcolor{gray}{CVPR'24}\\
    {FinePOSE} \cite{xu2024finepose} & 243 & CPN &  \underline{31.9} & \underline{25.0} & GT & \underline{16.7} & \underline{12.7} & \textcolor{gray}{CVPR'24}\\
    \midrule
    \textbf{{\proposed}} (Ours) & 243 & CPN &  \textbf{29.5} & \textbf{23.4} & GT & \textbf{15.9} & \textbf{12.2} & \\
    & & & \textcolor{blue}{(-2.4)} &  \textcolor{blue}{(-1.6)} & & \textcolor{blue}{(-0.8)} & \textcolor{blue}{(-0.5)} & \\
    \bottomrule
    \end{tabular}
}
\label{tab:SOTA-H3.6M}
\end{table*}

\begin{table*}[t]
\caption{\textbf{Quantitative comparison with the state-of-the-art 3D HPE methods on the Human3.6M dataset using 2D keypoint detectors to generate the inputs.} \textit{Dir.}, \textit{Disc.}, $\ldots$ , and \textit{Walk} correspond to 15 action classes. \textit{Avg} indicates the average MPJPE among 15 action classes. The best and second-best results are highlighted in \textbf{bold} and \underline{underlined} formats.}
\adjustbox{width=\linewidth}
  {
  \setlength{\tabcolsep}{4.3pt}
    \begin{tabular}{lcccccccccccccccl}
    \toprule
    \multirow{2}{*}{Method / mPJPE~$\textcolor{blue}{\downarrow}$} 
    & \multicolumn{16}{c}{Human3.6M (DET)} \\ 
    \cmidrule(lr){2-17}
    & Dir. & Disc. & Eat & Greet & Phone & Photo & Pose & Pur. & Sit & SitD. & Smoke & Wait & WalkD. & Walk & WalkT. & Avg \\ 
    \midrule
    VideoPose3D~\cite{videopose3d} & 45.2 & 46.7 & 43.3 & 45.6 & 48.1 & 55.1 & 44.6 & 44.3 & 57.3 & 65.8 & 47.1 & 44.0 & 49.0 & 32.8 & 33.9 & 46.8\\
    SRNet~\cite{SRNet} & 46.6 & 47.1 & 43.9 & 41.6 & 45.8 & 49.6 & 46.5 & 40.0 & 53.4 & 61.1 & 46.1 & 42.6 & 43.1 & 31.5 & 32.6 & 44.8\\
    RIE~\cite{rie} & 40.8 & 44.5 & 41.4 & 42.7 & 46.3 & 55.6 & 41.8 & 41.9 & 53.7 & 60.8 & 45.0 & 41.5 & 44.8 & 30.8 & 31.9 & 44.3\\
    Anatomy~\cite{anatomy} & 41.4 & 43.5 & 40.1 & 42.9 & 46.6 & 51.9 & 41.7 & 42.3 & 53.9 & 60.2 & 45.4 & 41.7 & 46.0 & 31.5 & 32.7 & 44.1\\
    P-STMO~\cite{pstmo} & 38.9 & 42.7 & 40.4 & 41.1 & 45.6 & 49.7 & 40.9 & 39.9 & 55.5 & 59.4 & 44.9 & 42.2 & 42.7 & 29.4 & 29.4 & 42.8 \\
    MixSTE~\cite{mixste} & 36.7 & 39.0 & 36.5 & 39.4 & 40.2 & 44.9 & 39.8 & 36.9 & 47.9 & 54.8 & 39.6 & 37.8 & 39.3 & 29.7 & 30.6 & 39.8\\
    MHFormer~\cite{mhformer} & 39.2 & 43.1 & 40.1 & 40.9 & 44.9 & 51.2 & 40.6 & 41.3 & 53.5 & 60.3 & 43.7 & 41.1 & 43.8 & 29.8 & 30.6 & 43.0\\
    Diffpose~\cite{diffpose_iccv} & 33.2 & 36.6 & 33.0 & 35.6 & 37.6 & 45.1 & 35.7 & 35.5 & 46.4 & 49.9 & 37.3 & 35.6 & 36.5 & 24.4 & 24.1 & 36.9\\
    GLA-GCN~\cite{glagcn} & 41.3 & 44.3 & 40.8 & 41.8 & 45.9 & 54.1 & 42.1 & 41.5 & 57.8 & 62.9 & 45.0 & 42.8 & 45.9 & 29.4 & 29.9 & 44.4\\
    ActionPrompt~\cite{actionprompt} & 37.7 & 40.2 & 39.8 & 40.6 & 43.1 & 48.0 & 38.8 & 38.9 & 50.8 & 63.2 & 42.0 & 40.0 & 42.0 & 30.5 & 31.6 & 41.8\\
    NC-RetNet~\cite{ncretnet} & 36.9 & 40.1 & 38.7 & 38.3 & 42.9 & 48.6 & 38.2 & 40.0 & 52.5 & 55.4 & 42.3 & 38.7 & 39.7 & 26.2 & 27.8 & 40.4\\
    MotionBERT~\cite{motionbert2022} & 36.1 & 37.5 & 35.8 & {32.1} & 40.3 & 46.3 & 36.1 & 35.3 & 46.9 & 53.9 & 39.5 & 36.3 & 35.8 & 25.1 & 25.3 & 37.5 \\
    D3DP~\cite{d3dp} & 33.0 & 34.8 & 31.7 & 33.1 & 37.5 & 43.7 & 34.8 & 33.6 & 45.7 & 47.8 & 37.0 & 35.0 & 35.0 &  24.3 &  24.1 &  35.4\\
    KTPFormer \cite{ktpformer} & \underline{30.1} & {32.1} & {29.1} & {30.6} & {35.4} & {39.3} & {32.8} & {30.9} & {43.1} & {45.5} & {34.7} & {33.2} & {32.7} & \underline{22.1} & {23.0} & {33.0}\\
    FinePOSE \cite{xu2024finepose} & {31.4} & \underline{31.5} & \underline{28.8} & \underline{29.7} & \underline{34.3} & \underline{36.5} & \underline{29.2} & \underline{30.0} & \underline{42.0} & \underline{42.5} & \underline{33.3} & \underline{31.9} & \underline{31.4} & {22.6} & \underline{22.7} & \underline{31.9}\\
    \midrule
   \textbf{{\proposed}}~(Ours)& \textbf{26.5} & \textbf{28.9} & \textbf{26.3} & \textbf{27.0} & \textbf{31.6} & \textbf{34.9} & \textbf{27.1} & \textbf{28.0} & \textbf{38.8} & \textbf{41.1} & \textbf{30.3} & \textbf{29.4} & \textbf{29.5} & \textbf{21.6} & \textbf{22.0} & \textbf{29.5}\\    
   & \textcolor{blue}{(-3.6)} & \textcolor{blue}{(-2.6)} & \textcolor{blue}{(-2.5)} & \textcolor{blue}{(-2.3)} & \textcolor{blue}{(-2.7)} & \textcolor{blue}{(-1.6)} & \textcolor{blue}{(-2.1)} & \textcolor{blue}{(-2.0)} & \textcolor{blue}{(-3.2)} & \textcolor{blue}{(-1.4)} & \textcolor{blue}{(-2.0)} & \textcolor{blue}{(-2.5)} & \textcolor{blue}{(-1.9)} & \textcolor{blue}{(-0.5)} & \textcolor{blue}{(-0.7)} & \textcolor{blue}{(-2.4)} \\
    \bottomrule
    \end{tabular}
}
\label{tab:quantitive-H3.6M-det}
\end{table*}

\section{Experiments}\label{sec:exp}

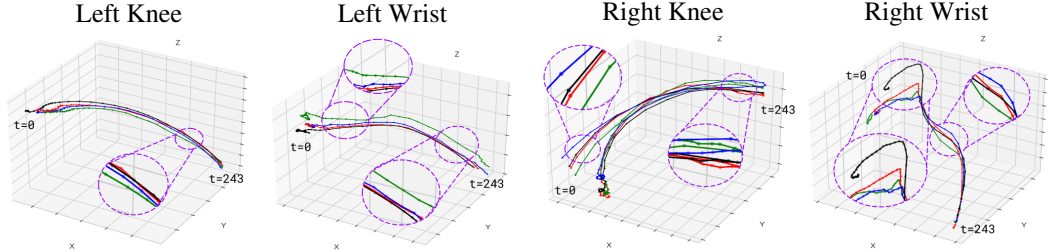
\begin{figure}[t]
{\centering
  \begin{tikzpicture}
    \node at (0,0) {\includegraphics[width=\linewidth]{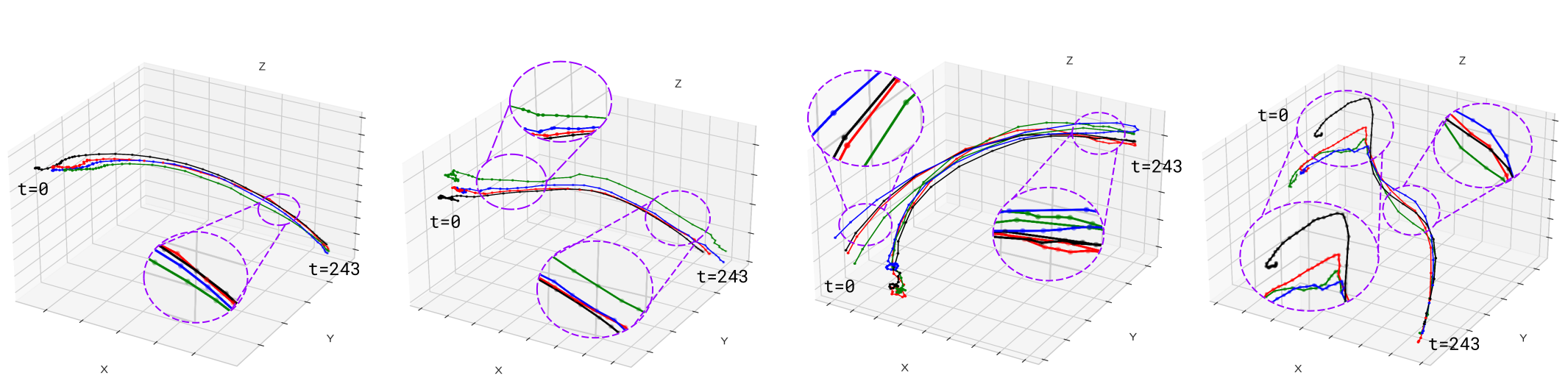}};
    \node at (-5.3,1.5) {Left Knee};
    \node at (-1.8,1.5) {Left Wrist};
    \node at (1.8,1.5) {Right Knee};
    \node at (5.3,1.5) {Right Wrist};
  \end{tikzpicture}
\vspace{-20px}
\caption{\textbf{Comparison of trajectories on the Human3.6M dataset for a sitting action.} The \textbf{black} trajectory represents the groundtruth, while the \textcolor{blue}{\textbf{blue}}, \textcolor{myGreen}{\textbf{green}}, and \textcolor{red}{\textbf{red}} trajectories correspond to the predictions from FinePOSE \cite{xu2024finepose}, KTPFormer \cite{ktpformer}, and {\proposed}, respectively.}
\label{fig:arm_traj}
}
\end{figure}

\begin{table*}[t]
\centering
\begin{minipage}{0.53\linewidth}
\centering
\caption{\small \textbf{Quantitative comparison with the SOTA 3D HPE methods on the MPI-INF-3DHP dataset using groundtruth 2D keypoints as inputs.} $N$: the number of input frames. The best and second-best results are highlighted in \textbf{bold} and \underline{underlined} formats.} \label{tab:SOTA-MPI}
\adjustbox{width=\linewidth}{
\setlength{\tabcolsep}{5pt}
\begin{tabular}{llccc}
\toprule
\multirow{2}{*}{Method} & \multirow{2}{*}{$N$} & \multicolumn{3}{c}{MPI-INF-3DHP} \\ 
\cmidrule(lr){3-5}
& & PCK$\textcolor{red}{\uparrow}$ & AUC$\textcolor{red}{\uparrow}$ & mPJPE$\textcolor{blue}{\downarrow}$ \\ 
\midrule
VideoPose3D~\cite{videopose3d} & 81 & 86.0 & 51.9 & 84.0 \\
MixSTE~\cite{mixste} & 27 & 94.4 & 66.5 & 54.9 \\
PoseFormerV2~\cite{poseformerv2} & 81 & 97.9 & 78.8 & 27.8 \\
MHFormer~\cite{mhformer} & 9 & 93.8 & 63.3 & 58.0 \\
Diffpose~\cite{diffpose_iccv} & 81 & 98.0 & 75.9 & 29.1 \\
D3DP~\cite{d3dp} & 243 & 98.0 & 79.1 & 28.1 \\
FinePOSE~\cite{xu2024finepose} & 243 & \underline{98.9} & 80.0 & 26.2 \\
KTPFormer~\cite{ktpformer} & 27 & \underline{98.9} & \underline{84.4} & \underline{19.2} \\
\midrule
\textbf{{\proposed} (Ours)} & 81 & \textbf{99.1} & \textbf{84.5} & \textbf{18.9} \\
& & \textcolor{red}{(0.2)} & \textcolor{red}{(0.1)} & \textcolor{blue}{(-0.3)} \\
\bottomrule
\end{tabular}
}
\end{minipage}
\hfill
\begin{minipage}{0.45\linewidth}
\centering
\caption{\small \textbf{Quantitative comparison with the SOTA 3D HPE methods on the MLBPitchDB dataset using 2D keypoints from groundtruth and a detector (ViTPose) as inputs.} $N$: the number of input frames. The best and second-best results are highlighted in \textbf{bold} and \underline{underlined} formats.} \label{tab:SOTA-MLBPitchDB}
\adjustbox{width=\linewidth}{
\setlength{\tabcolsep}{5pt}
\begin{tabular}{llcc}
\toprule
\multirow{2}{*}{Method} & \multirow{2}{*}{$N$} & \multicolumn{2}{c}{2D Keypoints} \\ 
\cmidrule(lr){3-4}
& & GT$\textcolor{blue}{\downarrow}$ & ViTPose$\textcolor{blue}{\downarrow}$ \\ 
\midrule
VideoPose3D~\cite{videopose3d} & 351 & 40.2 & 75.8 \\
MHFormer~\cite{mhformer} & 351 & 34.8 & 67.5 \\
D3DP~\cite{d3dp} & 243 & 26.6 & 61.7 \\
FinePOSE~\cite{xu2024finepose} & 243 & \underline{23.9} & \underline{57.3} \\
\midrule
\textbf{{\proposed} (Ours)} & 243 & \textbf{21.8} & \textbf{53.5} \\
& & \textcolor{blue}{(-2.1)} & \textcolor{blue}{(-3.8)} \\
\bottomrule
\end{tabular}
}
\end{minipage}
\end{table*}

\subsection{Quantitative Results}

\textbf{Human3.6M.} Table \ref{tab:SOTA-H3.6M} and \ref{tab:quantitive-H3.6M-det} presents a quantitative comparison of {\proposed} with various state-of-the-art (SOTA) 3D HPE techniques on the Human3.6M dataset. We conducted experiments using two different types of inputs following the literature of 3D HPE: 2D poses from a keypoint detector \cite{cpn} and groundtruth 2D poses. The results, as shown in Table \ref{tab:SOTA-H3.6M}, demonstrate the superior performance of {\proposed} on both input settings. {\proposed} outperforms the previous SOTA method, FinePOSE \cite{xu2024finepose}, by 7.5\% and 6.4\% in terms of mPJPE and P-mPJPE, respectively, when using a 2D keypoint detector. When using groundtruth 2D keypoints, {\proposed} still outperforms with improvements of 4.8\% and 3.9\% for mPJPE and P-mPJPE, respectively. Table \ref{tab:quantitive-H3.6M-det} further breaks down the results on the 15 action classes used in the Human3.6M dataset. Notably, {\proposed} achieves significant improvements in action categories such as "Directions", "Sit", and "Phone", with reductions in mPJPE of 11.9\%, 7.6\%, and 2.7\%, respectively. 

\noindent \textbf{MPI-INF-3DHP.} Table \ref{tab:SOTA-MPI} reports a comparison between {\proposed} and various SOTA techniques on the MPI-INF-3DHP dataset. The results highlight that {\proposed} achieves SOTA performance across key metrics, including PCK, AUC, and mPJPE. These improvements show the strong generalization capability of {\proposed}, particularly in challenging outdoor scenes consisting of variations in pose, lighting, and background conditions. 

\noindent \textbf{MLBPitchDB.} To evaluate the \textbf{generalizability} of {\proposed}, we use the MLBPitchDB dataset \cite{bright2023mitigating}, which consists of high inherent motion blur and occlusions. Table \ref{tab:SOTA-MLBPitchDB} demonstrates the SOTA performance of {\proposed} despite the challenging poses and the noisy 2D pose quality. Our method demonstrates a significant improvement of 8.7\% and 6.6\% with groundtruth 2D poses and 2D poses from a detector \cite{xu2022vitpose}, respectively. This highlights the effectiveness of {\proposed} in handling complex scenarios where motion blur and occlusion affect the quality of pose estimation.

\subsection{Qualitative Analysis}

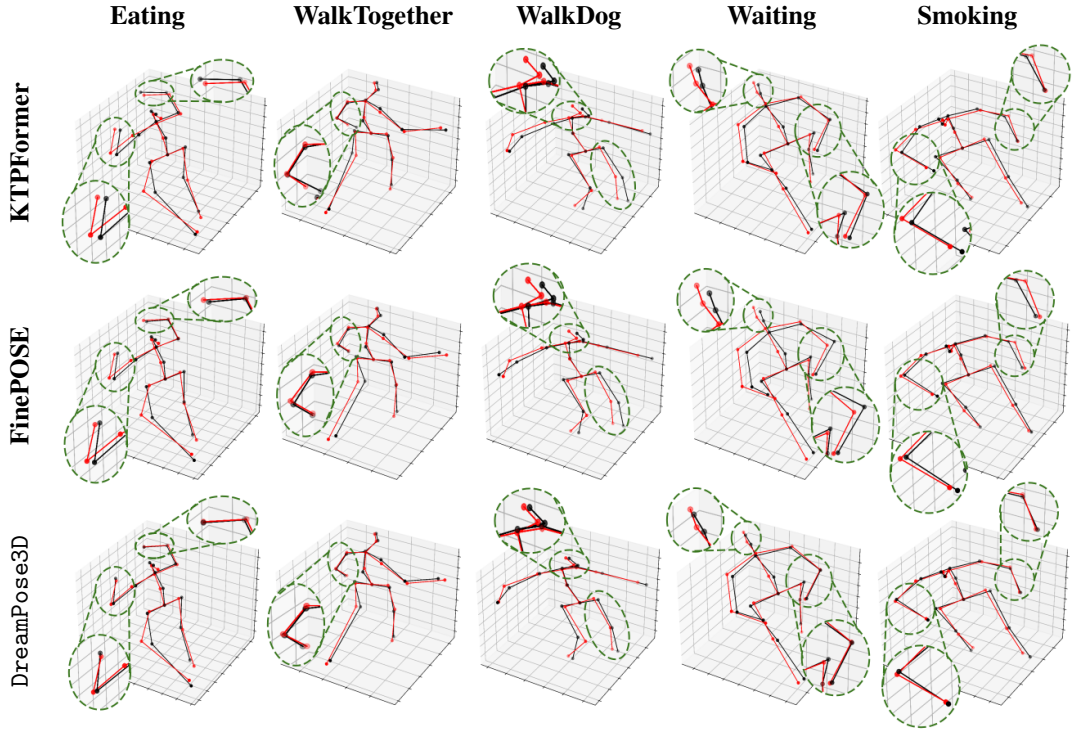
\begin{figure*}[t]
{\centering
  \begin{tikzpicture}
    \node at (0,0) {\includegraphics[width=\linewidth]{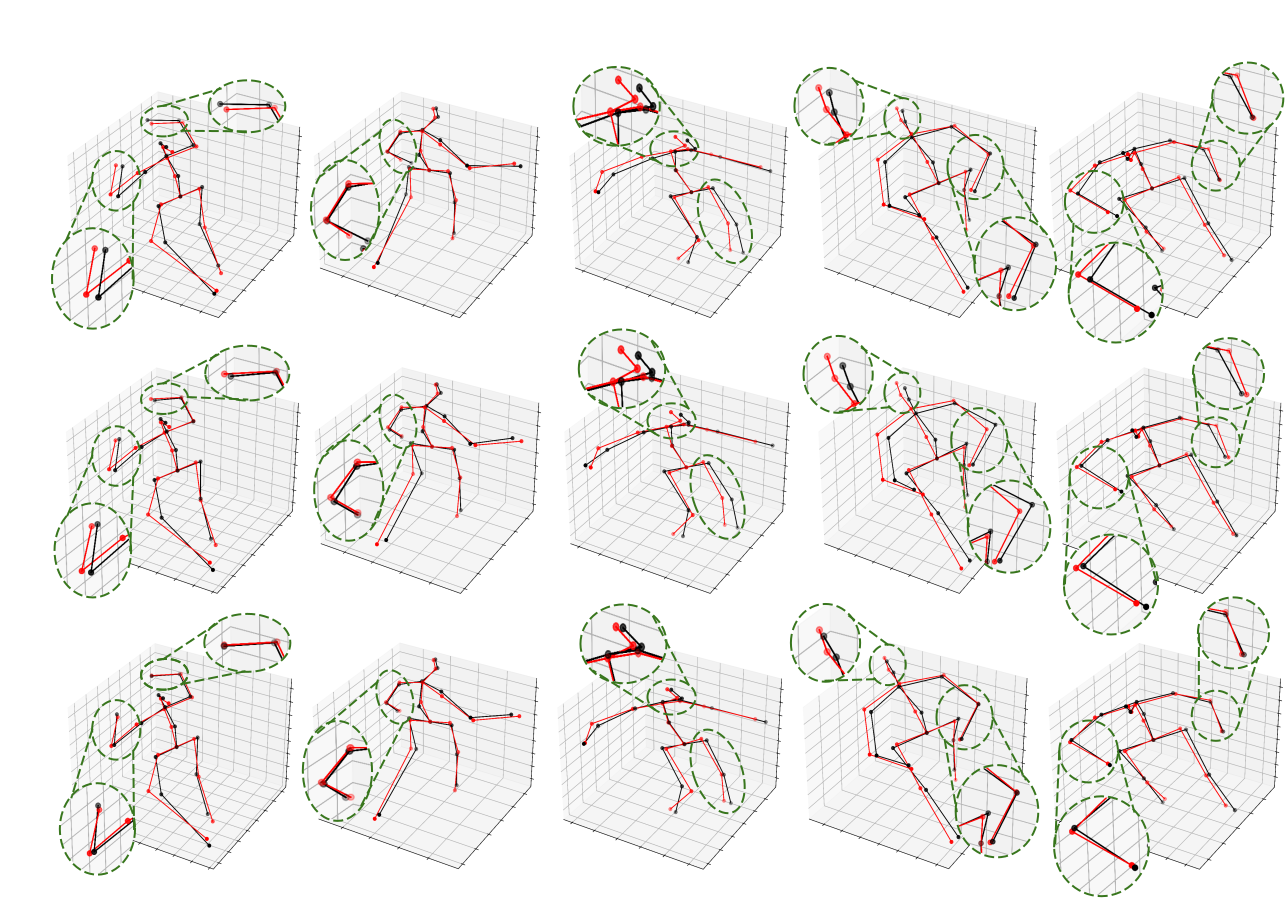}};
    \node at (-5.3,4.6) {\textbf{Eating}};
    \node at (-2.3,4.6) {\textbf{WalkTogether}};
    \node at (0.3,4.6) {\textbf{WalkDog}};
    \node at (3,4.6) {\textbf{Waiting}};
    \node at (5.6,4.6) {\textbf{Smoking}};
    
    \node[rotate=90] at (-7,2.8) {\textbf{KTPFormer}};
    \node[rotate=90] at (-7,-0.3) {\textbf{FinePOSE}};
    \node[rotate=90] at (-7,-3.3) {\textbf{{\proposed}}};
  \end{tikzpicture}
\vspace{-20px}
\caption{\textbf{Qualitative comparison of {\proposed} with FinePOSE \cite{xu2024finepose} and KTPFormer \cite{ktpformer} on the Human3.6M dataset}. The \textcolor{red}{\textbf{red}} skeleton represents the groundtruth pose, while the \textbf{black} skeleton shows {\proposed}’s prediction. \textcolor{myGreen}{\textbf{Green}} dashed lines highlight regions where {\proposed} demonstrates superior accuracy over the other methods.} \label{fig:qualitative}
}
\end{figure*}

The qualitative results of {\proposed} are demonstrated in Figure \ref{fig:qualitative}. We compare our method with two SOTA techniques, FinePOSE \cite{xu2024finepose} and KTPFormer \cite{ktpformer}. All estimated poses are overlaid alongside the groundtruth 3D poses for different actions. It can be observed that, especially for actions like "Eating" and "Waiting", {\proposed} produces poses that are more closely aligned with the groundtruth. This shows the ability of {\proposed} to model robust kinematic joint relationships from the SRE block, resulting in more accurate pose reconstruction, whereas prior methods show greater deviation from the groundtruth.

\begin{wrapfigure}{r}{0.5\linewidth}
    \centering
    \vspace{-17px}
    \begin{tikzpicture}
        \node at (0,0) {\includegraphics[width=\linewidth]{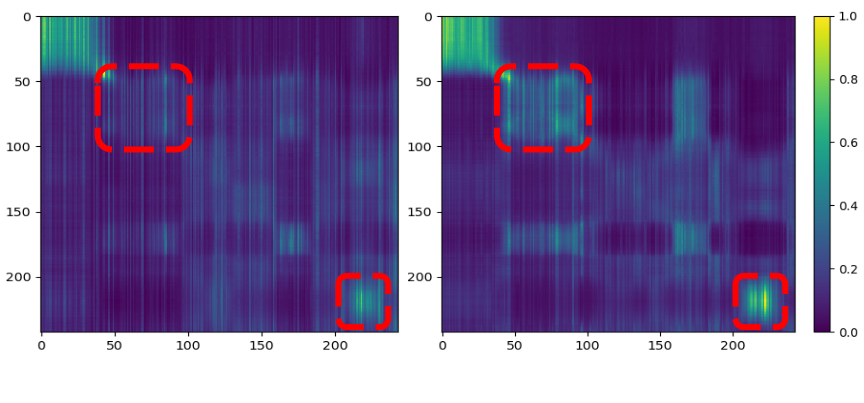}};
        \node at (-1.7,-1.4) {(a) FinePOSE};
        \node at (1.5,-1.4) {(b) {\proposed}};
    \end{tikzpicture}
    \vspace{-20px}
    \caption{\textbf{Comparison of the attention maps between ours and FinePOSE~\cite{xu2024finepose}.} The x-axis corresponds to queries and the y-axis to predicted outputs. Lighter color indicates stronger attention.} 
    \label{fig:att_map}
\end{wrapfigure}

Figure~\ref{fig:arm_traj} illustrates the consistency of the trajectories for four different joints of a human performing the "Sitting" action. Compared to prior methods, {\proposed} produces smoother and plausible motion paths, highlighting its ability to generate temporally coherent 3D poses through hallucination. Furthermore, Figure~\ref{fig:att_map} shows the temporal attention map for a sequence, where the x-axis corresponds to the query of 243 frames and the y-axis indicates the attention output. The highlighted regions demonstrate high attention-weight concentration, contributing to the ability of {\proposed} to hallucinate coherent poses and intent-conditioned denoising. 

\begin{table*}[t]
\centering
\begin{minipage}{0.49\linewidth}
\centering
\caption{\small \textbf{Ablation study on different configurations of {\proposed} on the Human3.6M dataset using 2D keypoint detectors as inputs.} All configurations use the denoiser backbone~\cite{mixste}.}
\adjustbox{width=\linewidth}{
\setlength{\tabcolsep}{6pt}
\begin{tabular}{lccccc}
\toprule
\multirow{2}{*}{Method} & \multicolumn{3}{c}{Module} & \multirow{2}{*}{mPJPE~$\textcolor{blue}{\downarrow}$} & \multirow{2}{*}{P-mPJPE~$\textcolor{blue}{\downarrow}$} \\
\cmidrule(lr){2-4}
& SRE & APL & HPD & & \\
\midrule
Baseline  & & & & 37.4 & 30.7 \\
w/o APL   & \checkmark & & \checkmark & 31.9 & 24.9 \\
w/o SRE   & & \checkmark & \checkmark & 32.2 & 24.8 \\
w/o HPD   & \checkmark & \checkmark & & 30.1 & 24.2 \\
\midrule
\textbf{{\proposed} (Ours)} & \checkmark & \checkmark & \checkmark & \textbf{29.5} & \textbf{23.4} \\
& & & & \textcolor{blue}{(-0.6)} & \textcolor{blue}{(-0.8)} \\
\bottomrule
\end{tabular}
}
\label{tab:ablation-modules}
\end{minipage}
\hfill
\begin{minipage}{0.49\linewidth}
\centering
\caption{\small \textbf{Impact of objective functions in {\proposed} on the Human3.6M dataset using 2D keypoint detectors as inputs.} Network fails to converge without $\mathcal{L}'_{3D}$ due to lack of pose regularization.}
\adjustbox{width=\linewidth}{
\setlength{\tabcolsep}{6pt}
\begin{tabular}{lccccc}
\toprule
\multirow{2}{*}{Method} & \multicolumn{3}{c}{Module} & \multirow{2}{*}{mPJPE~$\textcolor{blue}{\downarrow}$} & \multirow{2}{*}{P-mPJPE~$\textcolor{blue}{\downarrow}$} \\
\cmidrule(lr){2-4}
& $\mathcal{L}'_{3D}$ & $\mathcal{L}_{act}$ & $\mathcal{L}_{BL}$ & & \\
\midrule
w/ $\mathcal{L}'_{3D}$                     & \checkmark & & & 31.9 & 25.6 \\
w/ $\mathcal{L}'_{3D} + \mathcal{L}_{act}$ & \checkmark & \checkmark & & 31.1 & 24.8 \\
w/ $\mathcal{L}'_{3D} + \mathcal{L}_{BL}$  & \checkmark & & \checkmark & 30.4 & 24.0 \\
\midrule
\textbf{{\proposed} (Ours)} & \checkmark & \checkmark & \checkmark & \textbf{29.5} & \textbf{23.4} \\
& & & & \textcolor{blue}{(-0.9)} & \textcolor{blue}{(-0.6)} \\
\bottomrule
\end{tabular}
}
\label{tab:ablation-loss}
\end{minipage}
\end{table*}

\subsection{Ablation Study}

\textbf{Effect of different blocks on \textbf{\proposed}.} To quantify the influence of individual blocks on {\proposed}, we conduct an ablation as presented in Table \ref{tab:ablation-modules}. Starting with a baseline model consisting solely of a backbone denoiser block \cite{mixste}, we incorporated each of the proposed blocks. The final model, incorporating all components, achieved substantial performance gains of 7.9\% and 7.3\% in terms of mPJPE and P-mPJPE, respectively, compared to the denoiser-only baseline model. These results highlight the significance of intent, hallucinating 3D poses, and kinematic representation modeling for the 3D HPE task. 

\noindent \textbf{Effect of different objective functions on \textbf{\proposed}.} Table \ref{tab:ablation-loss} evaluates the influence of the bone length and prompt regularization on {\proposed} performance. Incorporating these losses into the objective function yields improvements of 2.4\% and 2.2\% in mPJPE and P-mPJPE, respectively. These results show the importance of explicit regularization of prompts and bone length modeling.


%% file: sec/5_conclusion.tex
\section{Conclusion}\label{sec:discussion}

In this work, we introduced {\proposed}, a novel framework for 3D human pose estimation that mimics human-like reasoning by inferring motion intent and hallucinating plausible 3D pose trajectories. Our method integrates the power of language models to produce action-aware information, which conditions a transformer-based denoiser to predict 3D hallucinatory poses. Furthermore, novel ways to incorporate kinematic joint information into the attention mechanism are explored for the denoiser. Experimentation on three datasets demonstrates the efficacy of our model in capturing robust spatial and temporal information by achieving SOTA performance in challenging scenarios through its intent-conditioned hallucination approach.

%% file: sec/X_suppl.tex

\newpage
\appendix
\section*{Supplementary Material}

The supplementary material is organized as follows:

\begin{description}[leftmargin=5em, labelsep=1em]
    \item[Section \ref{app:diff}] Additional Formulation of Diffusion Models
    \item[Section \ref{app:implementation}] Implementation Details
    \item[Section \ref{app:datasets}] Datasets
    \item[Section \ref{app:eval}] Evaluation Metrics
    \item[Section \ref{app:ablation}] More Ablation Study
    \item[Section \ref{app:qual}] More Qualitative Results
    \item[Section \ref{app:limitations}] Limitations and Future Works
    \item[Section \ref{app:impact}] Broader Impact Statement
    \item[Section \ref{app:findings}] Summary of Experimental Findings
\end{description}

\section{Additional Formulation of Diffusion Models} \label{app:diff}

Our work is built over DDPMs \cite{ddpm}, which are trained to reverse a diffusion process that gradually adds Gaussian noise to the training data. This can be formulated as a two-stage process:

\noindent \textbf{Forward process.} The diffusion process $q$ takes in as input a sequence of groundtruth 3D pose $\textbf{Y}_0 \in \mathbb{R}^{N \times J \times 3}$ and a sampled time step $t \in [0, T]$, where $T$ is the maximum number of diffusion steps. Then the input is diffused by adding Gaussian noise $\epsilon \sim \mathcal{N}(0, \textbf{I})$ at each step to generate $\textbf{Y}_t (t \rightarrow T)$. This can be mathematically written as

\begin{equation}
    q(\textbf{Y}_t | \textbf{Y}_0) = \sqrt{\Bar{\alpha}} \textbf{Y}_0 + \epsilon \sqrt{1- \Bar{\alpha}} 
\end{equation}

\noindent where $\Bar{\alpha}$ is a constant dependent on the variance schedule.

\noindent \textbf{Reverse process.} The reverse process aims in reconstructing the uncontaminated 3D poses $\Bar{\textbf{Y}}_0$ using a denoiser $\mathcal{D}$ from the noised 3D poses $\textbf{Y}_t$ from the forward process:

\begin{equation}
    \hat{\textbf{Y}}_0 = \mathcal{D}(\textbf{Y}_t, \textbf{X}, t)  
\end{equation}

\noindent where the denoiser $\mathcal{D}$ takes as input the noised 3D pose sequence $\textbf{Y}_t$, given the 2D pose sequence $\textbf{X}$ and the time step to reconstruct the denoised output 3D pose sequence $\hat{\textbf{Y}}_0$.

\section{Implementation Details} \label{app:implementation}

{\proposed} is implemented in PyTorch \cite{pytorch} and optimized using the AdamW optimizer \cite{adamw} with a learning rate of $1 \times 10^{-5}$ and a weight decay of $1 \times 10^{-4}$ per epoch. The model is trained on three A6000 GPUs with a batch size of 4 for 100 epochs, which takes approximately two days. During training, we freeze the pretrained CLIP model to avoid updating its weights. The denoiser backbone (SPD) is based on MixSTE \cite{mixste}, which incorporates $M=16$ stacks of spatial and temporal transformers with a channel size of 512. The SRE block utilizes MHSA with $L=6$.


\section{Datasets} \label{app:datasets}

\textbf{Human3.6M.} The Human3.6M dataset \cite{h36m} is the standard benchmark for 3D HPE in controlled indoor environments. It comprises over 3.6 million images capturing 15 daily activities performed by 11 subjects at a frame rate of 50 Hz. Following common practice \cite{xu2024finepose, d2ahmr}, we train on data from five subjects (S1, S5, S6, S7, S8) and evaluate on two subjects (S9, S11). Some action prompts used to train $MLP_p$ with Human3.6M dataset include "Sitting", "Phoning", "Greeting", "Discussion", etc. 

\noindent \textbf{MPI-INF-3DHP.} The MPI-INF-3DHP dataset \cite{mpi3dhp} is the largest publicly available 3D human pose dataset capturing diverse motions in both indoor and outdoor environments. Comprising over 1.3 million images, it surpasses the motion variety offered by the Human3.6M dataset. Some action prompts used to train $MLP_p$ with the MPI-INF-3DHP dataset include "Walking", "Running", "Crouching", "Bending", etc. 

\noindent \textbf{MLBPitchDB.} The MLBPitchDB dataset \cite{bright2023mitigating} comprises 30,000 images spanning 150 pitch sequences from diverse MLB games. Characterized by inherent blur and occlusions due to the rapid motion of baseball players captured at 30fps by broadcast cameras, this dataset rigorously evaluates {\proposed}'s capacity to generate accurate and robust pose predictions under challenging real-world conditions. The $MLP_p$ block was trained with action prompts including "Throwing" and "Hitting" based on the performed action by the baseball players. 

\section{Evaluation Metrics} \label{app:eval}

To evaluate the accuracy of {\proposed}, we compute the mean Per Joint Position Error (mPJPE) \cite{h36m} and Procrustes-aligned mPJPE (PA-mPJPE) \cite{pampjpe} measured in millimeters (mm). We report the mPJPE and PA-mPJPE between the predictions and groundtruth 3D poses. In addition, the Percentage of Correct Keypoint (PCK) with a threshold of 150 mm, and Area Under Curve (AUC) for a range of PCK thresholds are used as metrics for the MPI-INF-3DHP dataset. Following typical work on 3D human pose estimation \cite{d2ahmr, xu2024finepose, motionbert2022}, we report the metrics in Human3.6M validation data split and the test set of the MPI-INF-3DHP and MLBPitchDB datasets. 

\section{More Ablation Study} \label{app:ablation}

\begin{figure}[t]
    \centering
    \begin{subfigure}[b]{0.48\textwidth}
        \centering
        \begin{tikzpicture}
            \node at (0,0) {\includegraphics[width=\linewidth]{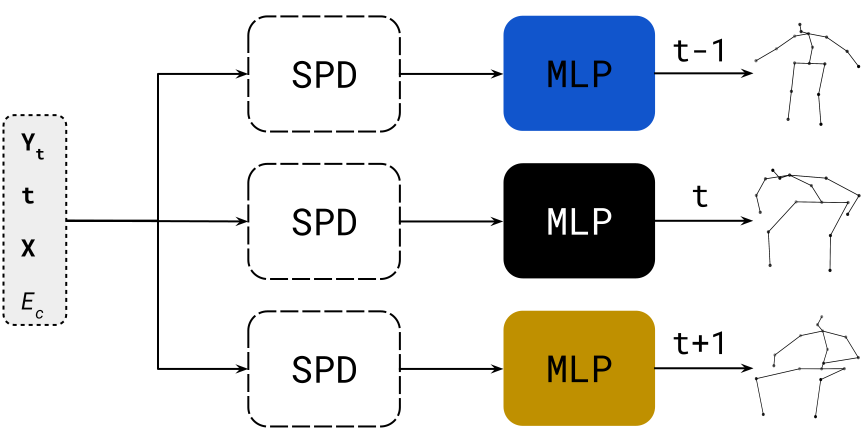}};
        \end{tikzpicture}
        \caption{Hallucinator-Free Design}
        \label{fig:config_a}
    \end{subfigure}
    \hfill
    \begin{subfigure}[b]{0.48\textwidth}
        \centering
        \begin{tikzpicture}
            \node at (0,0) {\includegraphics[width=\linewidth]{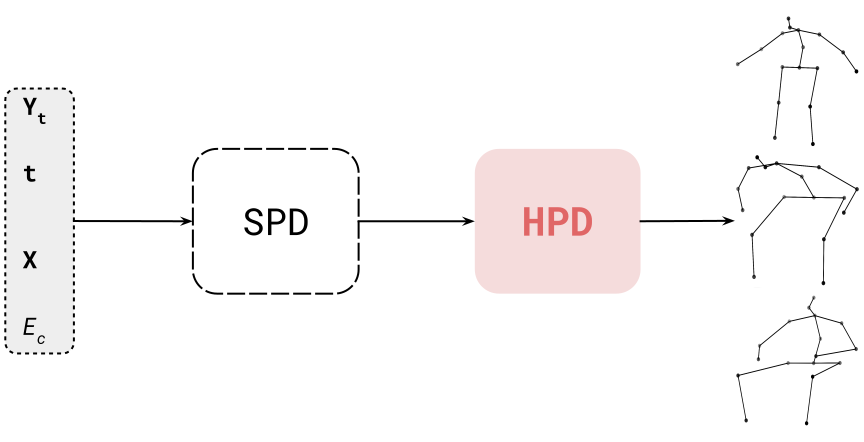}};
        \end{tikzpicture}
        \caption{Hallucinator Design}
        \label{fig:config_b}
    \end{subfigure}
    \hfill

    \caption{\textbf{Different module configurations for {\proposed}.} (a) illustrates alternative setup for the proposed {\proposed} module (b).}
    \label{fig:model_conf}
\end{figure}

\noindent \textbf{Effect of different model configurations on {\proposed}.} Table \ref{tab:pha_conf} presents an ablation study comparing an alternative design based on model simplicity. Hallucinator-free design, depicted in Figure \ref{fig:model_conf}(a), where the HPD block in {\proposed} (Figure \ref{fig:model_conf}(b)) is replaced with $n$ denoisers having shared weights to predict $n$ poses. Experimental results indicate that {\proposed} achieves significant improvements, with reductions of 3.9\% in mPJPE and 2.9\% in PA-mPJPE compared to the hallucinator-free design. Additionally, the hallucinator-free design incurred a 24.95-second increase in inference time compared to {\proposed}.

\begin{figure}[t]
\centering

\begin{minipage}[t]{0.48\textwidth}
\captionof{table}{\small \textbf{Quantitative comparison of different model configurations on {\proposed}.} $\text{CT}$: Computation Time.}
\adjustbox{width=\linewidth}{
\setlength{\tabcolsep}{5pt}
\begin{tabular}{lccc}
\toprule
 & mPJPE~$\textcolor{blue}{\downarrow}$ & P-mPJPE~$\textcolor{blue}{\downarrow}$ & CT~$\textcolor{blue}{\downarrow}$ \\
\midrule
Hallucinator-free design & 30.7 & 24.1 & 85.62 \\
\midrule
{\proposed} & \textbf{29.5} & \textbf{23.4} & 60.67 \\
& \textcolor{blue}{(-1.2)} & \textcolor{blue}{(-0.7)} & \textcolor{red}{(+24.95)} \\
\bottomrule
\end{tabular}
}
\label{tab:pha_conf}
\end{minipage}
\hfill
\begin{minipage}[t]{0.48\textwidth}
\captionof{table}{\small \textbf{Quantitative comparison of the conventional sampling strategy on {\proposed} during training.}}
\adjustbox{width=\linewidth}{
\setlength{\tabcolsep}{6pt}
\begin{tabularx}{\linewidth}{l *{2}{>{\centering\arraybackslash}X}} 
\toprule
Method & mPJPE~$\textcolor{blue}{\downarrow}$ & P-mPJPE~$\textcolor{blue}{\downarrow}$ \\
\midrule
Fixed Weights Sam. & 30.1 & 24.3 \\
Falloff Sam. & 30.0 & 24.1 \\
\midrule
Controlled Sam. & \textbf{29.5} & \textbf{23.4} \\
& \textcolor{blue}{(-0.5)} & \textcolor{blue}{(-0.7)} \\
\bottomrule
\end{tabularx}
}
\label{tab:ablation-sampling}
\end{minipage}

\end{figure}

\begin{figure}[t]
\centering

\begin{minipage}[t]{0.48\textwidth}
\captionof{table}{\small \textbf{Impact of the number of hallucinated poses in HPD of {\proposed} on Human3.6M dataset using 2D keypoint detectors as inputs during training.} $n$ denotes the sequence length/ number of output predictions.}
\adjustbox{width=\linewidth}{
\setlength{\tabcolsep}{6pt}
\begin{tabularx}{\linewidth}{l *{2}{>{\centering\arraybackslash}X}} 
\toprule
Method & mPJPE~$\textcolor{blue}{\downarrow}$ & P-mPJPE~$\textcolor{blue}{\downarrow}$ \\
\midrule
HPD ($n=1$) & 30.1 & 24.2 \\
HPD ($n=5$) & 29.9 & 23.7 \\
HPD ($n=7$) & 30.0 & 23.9 \\
\midrule
HPD ($n=3$) & \textbf{29.5} & \textbf{23.4} \\
& \textcolor{blue}{(-0.5)} & \textcolor{blue}{(-0.5)} \\
\bottomrule
\end{tabularx}
}
\label{tab:ablation-pha}
\end{minipage}
\hfill
\begin{minipage}[t]{0.48\textwidth}
\captionof{table}{\small \textbf{Impact of the quality of text prompts on {\proposed}}}
\adjustbox{width=\linewidth}{
\begin{tabular}{lcc}
\toprule
Method & mPJPE~$\textcolor{blue}{\downarrow}$ & P-mPJPE~$\textcolor{blue}{\downarrow}$ \\
\midrule
No CLIP module (APL) & 31.9 & 24.9\\ 
Random Prompts & \underline{31.2} & \underline{24.0}\\ 
\midrule
\textbf{{\proposed}} (Ours) & \textbf{29.5} & \textbf{23.4}\\ 
& \textcolor{blue}{(-0.7)} & \textcolor{blue}{(-0.6)} \\
\bottomrule
\end{tabular}
}
\label{tab:ablation2}
\end{minipage}

\end{figure}

\begin{figure*}
{\centering
  \begin{tikzpicture}
    \node at (0,0) {\includegraphics[width=\linewidth]{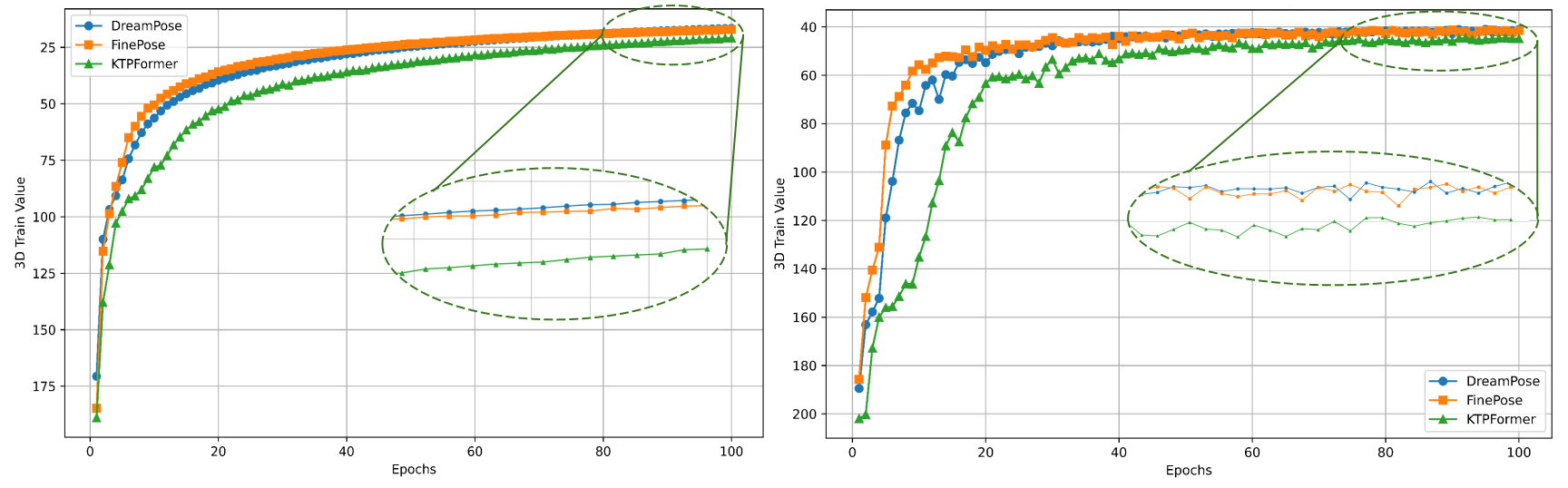}};
  \end{tikzpicture}
  \caption{\textbf{Training and validation curves} in Human3.6M dataset for FinePOSE\cite{xu2024finepose}, KTPFormer \cite{ktpformer} and {\proposed}.}
\label{fig:training_curves}
}
\end{figure*}

\begin{figure*}[t]
{\centering
  \begin{tikzpicture}
    \node at (0,0) {\includegraphics[width=\linewidth]{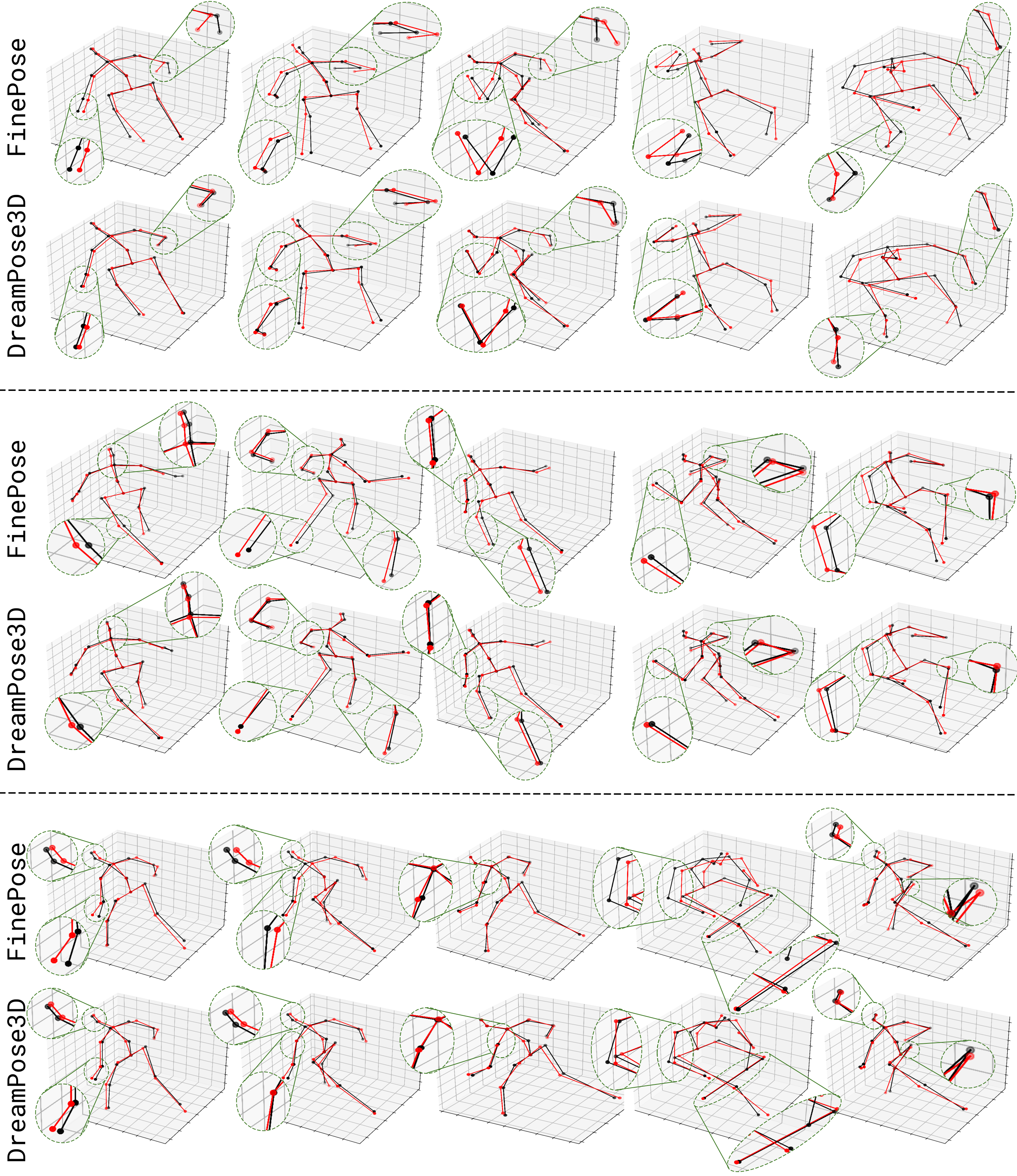}};
  \end{tikzpicture}
  \vspace{-25px}
  
\caption{\textbf{Qualitative comparison of {\proposed} with FinePOSE \cite{xu2024finepose}}. The \textcolor{red}{\textbf{red}} skeleton represents the groundtruth pose, while the \textbf{black} skeleton shows {\proposed}’s prediction. \textcolor{myGreen}{\textbf{Green}} dashed lines highlight regions where {\proposed} demonstrates superior accuracy over the other methods.} \label{fig:qualitative_sup}
}
\end{figure*} 

\paragraph{Sampling Strategy.} Table \ref{tab:ablation-sampling} presents an ablation study on different sampling strategies evaluated on the Human3.6M dataset. The strategies include: (1) \textbf{Fixed Weights Sampling}, where fixed weights, $\lambda_{3D}^{f+k}$, are assigned uniformly to all predictions throughout training; (2) \textbf{Falloff Sampling}, which assigns the highest weight to the center frame, with weights decreasing as the temporal distance from the center frame increases; and (3) \textbf{Controlled Sampling}, which restricts training to predict only the center frame ($n=1$) for the first 25 epochs. For the remaining epochs, predictions are reduced with a decay, represented as: $\lambda_{3D}^{f+k}=1/(1+|k|)$. Results indicate that the controlled sampling strategy achieves the best performance, highlighting the effectiveness of gradually increasing prediction difficulty during training with a decay factor.

\paragraph{Effect of different pose decoder configurations on {\proposed}.} Table \ref{tab:ablation-pha} presents an ablation study comparing the performance of the HPD block with varying numbers of hallucinated poses ($n$) during training. Experimental results demonstrate that the HPD consistently achieves optimal results with three poses ($n=3$) as output, resulting in 1.7\% and 2.1\% reductions in mPJPE and PA-mPJPE, respectively, when compared against HPD at $n=1$ during training. While in inference, $n$ is always set to 1 since the objective of hallucinating poses was only to improve the training coherence.

\paragraph{Impact of Text Prompts.} Table \ref{tab:ablation2} shows the impact the quality of text prompts have on {\proposed}. Interestingly, the models performed better with random prompts than when the APL block was entirely omitted. This suggests that the CLIP model positively influences the overall performance, regardless of the specific prompt used. However, employing relevant prompts yields the best results of 2.2\% and 2.5\% in terms of mPJPE and PA-mPJPE, respectively.

\section{More Qualitative Results} \label{app:qual}

In this section, we present more qualitative results of {\proposed}. Figure \ref{fig:qualitative_sup} depicts the visual comparison of {\proposed} with FinePOSE \cite{xu2024finepose} model which was the previous SOTA in 3DPE task. The highlighted regions depict the regions in which {\proposed} significantly outperforms FinePOSE in pose alignment with the groundtruth. In addition, Figure \ref{fig:training_curves} presents the training and validation curves of KTPFormer \cite{ktpformer}, FinePOSE \cite{xu2024finepose}, and the proposed {\proposed} method.  

\begin{figure}[t]
{\centering
  \begin{tikzpicture}
    \node at (0,0) {\includegraphics[width=0.7\linewidth]{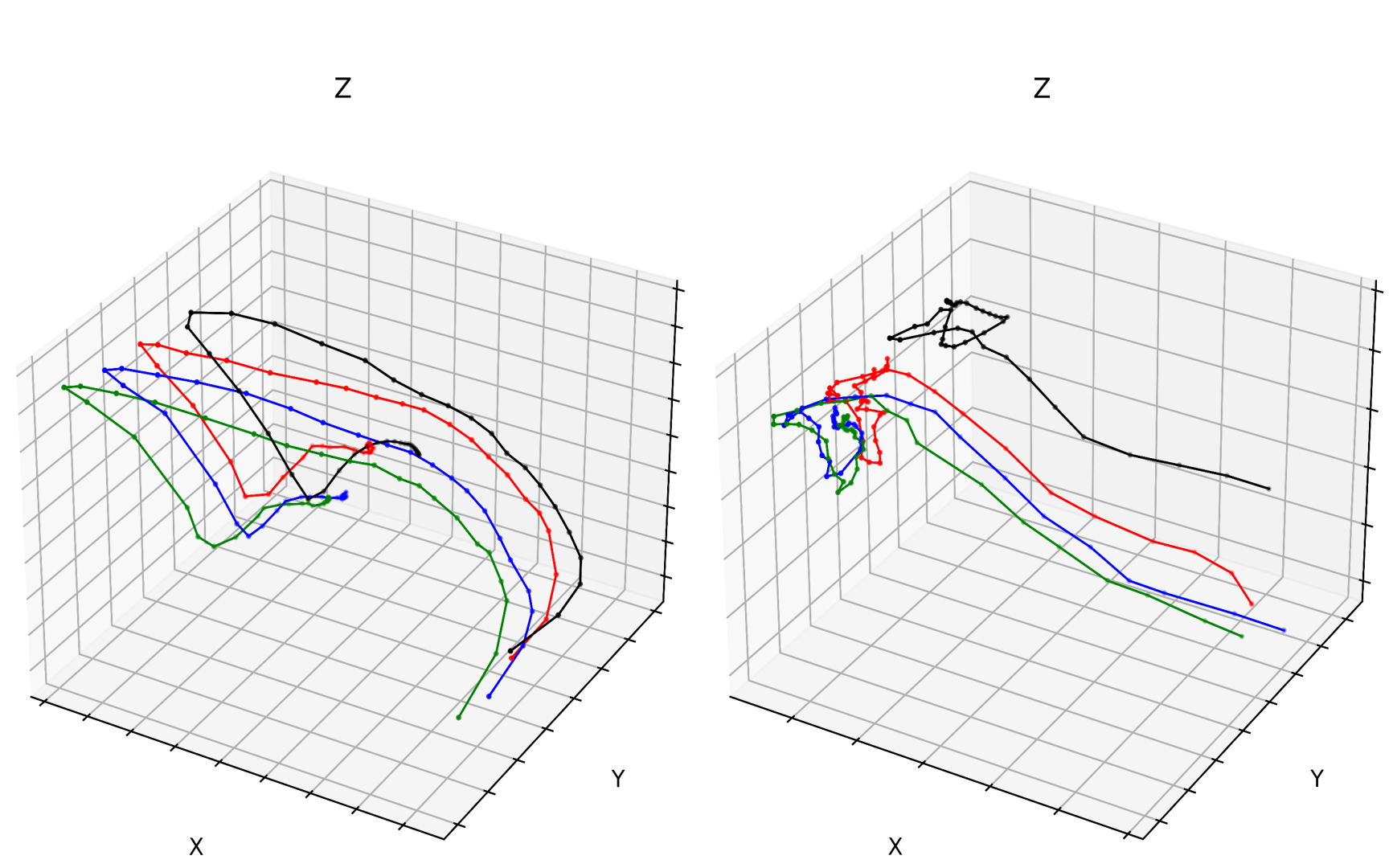}};
    \node at (-2.4, 3.0) {Left Foot};
    \node at (2.4, 3.0 ) {Left Hip};
  \end{tikzpicture}
  \caption{\textbf{Gap in Existing 3DPE Methods.} The \textbf{black} trajectory represents the groundtruth, while the \textcolor{blue}{\textbf{blue}}, \textcolor{myGreen}{\textbf{green}}, and \textcolor{red}{\textbf{red}} trajectories correspond to the predictions from FinePOSE \cite{xu2024finepose}, KTPFormer \cite{ktpformer}, and {\proposed}, respectively.}
\label{fig:limitation}
}
\end{figure}

\section{Limitations and Future Works} \label{app:limitations}

Inconsistencies in the predicted poses are observed, particularly in high-occlusion scenarios, including cases of self-occlusion by the target individual. This is illustrated in Figure \ref{fig:limitation}, where the trajectories of the left hip and left foot show noticeable misalignment compared to the groundtruth (black). Despite these misalignments, it is worth noting that {\proposed} demonstrates closer alignment to the groundtruth compared to prior SOTA methods, such as FinePOSE \cite{xu2024finepose} and KTPFormer \cite{ktpformer}.

For future research, we aim to explore the integration of physics-based priors to further enhance pose realism and incorporate image features for richer contextual understanding, potentially improving performance in highly unconstrained, in-the-wild settings. 


\section{Broader Impact Statement} \label{app:impact} 

By mimicking aspects of human cognition for human motion understanding, {\proposed} advances 3D pose estimation, benefiting applications in animation, sports analytics, augmented reality, and human-object interaction. The ability for such systems to act under ambiguous or noisy conditions makes it especially valuable for real-world deployment with no specific need for expensive sensors or controlled environmental setup.

More broadly, this work contributes to a growing trend of incorporating human-like reasoning and intent understanding into AI systems. Such domains can be adopted beyond just pose estimation, including assistive robotics, human-computer interaction, and embodied agents, where an implicit understanding of \textit{why} a person moves is necessary to find, rather than just \textit{how} they move.

While our work primarily focuses on technical improvements, we acknowledge that the deployment of pose estimation systems must carefully consider potential ethical concerns, such as misuse of surveillance or privacy violations. 

\section{Summary of Experimental Findings} \label{app:findings}

The experimental results systematically demonstrated the effectiveness of {\proposed}'s two cognitively inspired capabilities: \textit{intent inference} and \textit{hallucinative motion generation}. We evaluated their contributions through targeted ablations and visualizations across multiple datasets and configurations.

\textbf{Intention.}
Table 5 (from the main paper) shows that removing the APL block leads to a reduction of \textbf{7.5\%} in mPJPE, confirming the critical role of intent guidance. Table \ref{tab:ablation2} further demonstrates that the improvement is not solely due to CLIP embeddings, i.e., prompt structure and relevance significantly affect performance (\textbf{2.2}\% in mPJPE).

\textbf{Hallucination.}
Ablating the HPD block results in a performance drop of \textbf{2.0\%} in mPJPE as shown in Table 5 (from the main paper), highlighting its importance for temporal consistency. Figure 3 (from the main paper) visualizes smoother joint trajectories compared to prior SOTA, and Figure 4 (from the main paper) shows more concentrated attention weights than FinePose, supporting improved temporal reasoning. Tables \ref{tab:pha_conf} and \ref{tab:ablation-pha} also confirm that both the HPD design and the number of hallucinated outputs contribute positively to performance.

Together, these results confirm that {\proposed} benefits from the proposed modules. The model achieves state-of-the-art results on both Human3.6M and MPI-INF-3DHP, and shows strong generalization on the broadcast baseball dataset, handling occlusions, motion blur, resulting in ambiguous 2D inputs with greater robustness than prior approaches. This reinforces the value of incorporating intent and hallucination into temporally structured 3D pose estimation.